\pdfoutput=1

\documentclass[11pt]{article}

\usepackage{acl}

\usepackage{times}
\usepackage{latexsym}
\usepackage{booktabs} 
\usepackage{caption}  
\usepackage{amsmath}  
\usepackage{graphicx} 
\usepackage{multirow}
\usepackage{booktabs}
\usepackage{graphicx}  
\usepackage{caption}   
\usepackage{amsmath}
\usepackage{algorithm}
\usepackage{algorithmic}
\usepackage{amsfonts}
\usepackage{caption}

\usepackage[T1]{fontenc}

\usepackage[utf8]{inputenc}

\usepackage{microtype}

\usepackage{inconsolata}

\usepackage{graphicx}

\title{Layer Importance and Hallucination Analysis in Large Language Models via Enhanced Activation Variance-Sparsity}

\author{
  Zichen Song, Sitan Huang, Yuxin Wu, Zhongfeng Kang\thanks{* Corresponding author} \\
  Lanzhou University \\
  \texttt{songzch21@lzu.edu.cn, kangzf@lzu.edu.cn} \\
}

\begin{document}
\maketitle
\begin{abstract}
Evaluating the importance of different layers in large language models (LLMs) is crucial for optimizing model performance and interpretability. This paper first explores layer importance using the Activation Variance-Sparsity Score (AVSS), which combines normalized activation variance and sparsity to quantify each layer's contribution to overall model performance. By ranking layers based on AVSS and pruning the least impactful 25\%, our experiments on tasks such as question answering, language modeling, and sentiment classification show that over 90\% of the original performance is retained, highlighting potential redundancies in LLM architectures. Building on AVSS, we propose an enhanced version tailored to assess hallucination propensity across layers (EAVSS). This improved approach introduces Hallucination-Specific Activation Variance (HSAV) and Hallucination-Specific Sparsity (HSS) metrics, allowing precise identification of hallucination-prone layers. By incorporating contrastive learning on these layers, we effectively mitigate hallucination generation, contributing to more robust and efficient LLMs(The maximum performance improvement is 12\%). Our results on the NQ, SciQ, TriviaQA, TruthfulQA, and WikiQA datasets demonstrate the efficacy of this method, offering a comprehensive framework for both layer importance evaluation and hallucination mitigation in LLMs.
\end{abstract}

\section{Introduction}

Evaluating the importance of different layers in deep learning models is crucial for improving model efficiency, interpretability, and robustness. Identifying key layers allows for effective model compression and a more informed model design. Recently, large language models (LLMs) have shown remarkable capabilities across diverse applications, including question answering, language modeling, and sentiment analysis. However, there is limited research on the functional contributions of individual layers in LLMs, particularly from the perspective of activation variance and sparsity, which could reveal each layer's unique role in model performance and interpretability \cite{wang2024deepnet, xiong2020layernorm}. Moreover, studies specifically focusing on hallucination propensity based on layer activation patterns in LLMs remain largely unexplored, leaving a critical gap in understanding and mitigating layer-specific hallucination generation.

Previous works on layer importance have introduced several sophisticated methodologies. Saarela et al. \cite{saarela2021featureimportance} proposed Gradient-Based Importance Scores (GBIS), which assess layer importance by calculating the sensitivity of gradients relative to inputs, thereby reflecting model reliance on each layer’s activations. Zopf et al. \cite{bach2015pixelwise} introduced Layer-wise Relevance Propagation (LRP), analyzing information flow through the model and helping to understand the role of each layer in the model’s decision process. Additionally, Mencía et al. \cite{unknown2016sequential} developed Contextual Importance Measures (CIM), dynamically evaluating layer importance based on specific input conditions. While these methods offer valuable insights, they often fall short in capturing complex activation patterns and identifying redundancy in LLMs, particularly as model depth and size increase.

\begin{figure*}[ht]
    \centering
    \includegraphics[width=\textwidth]{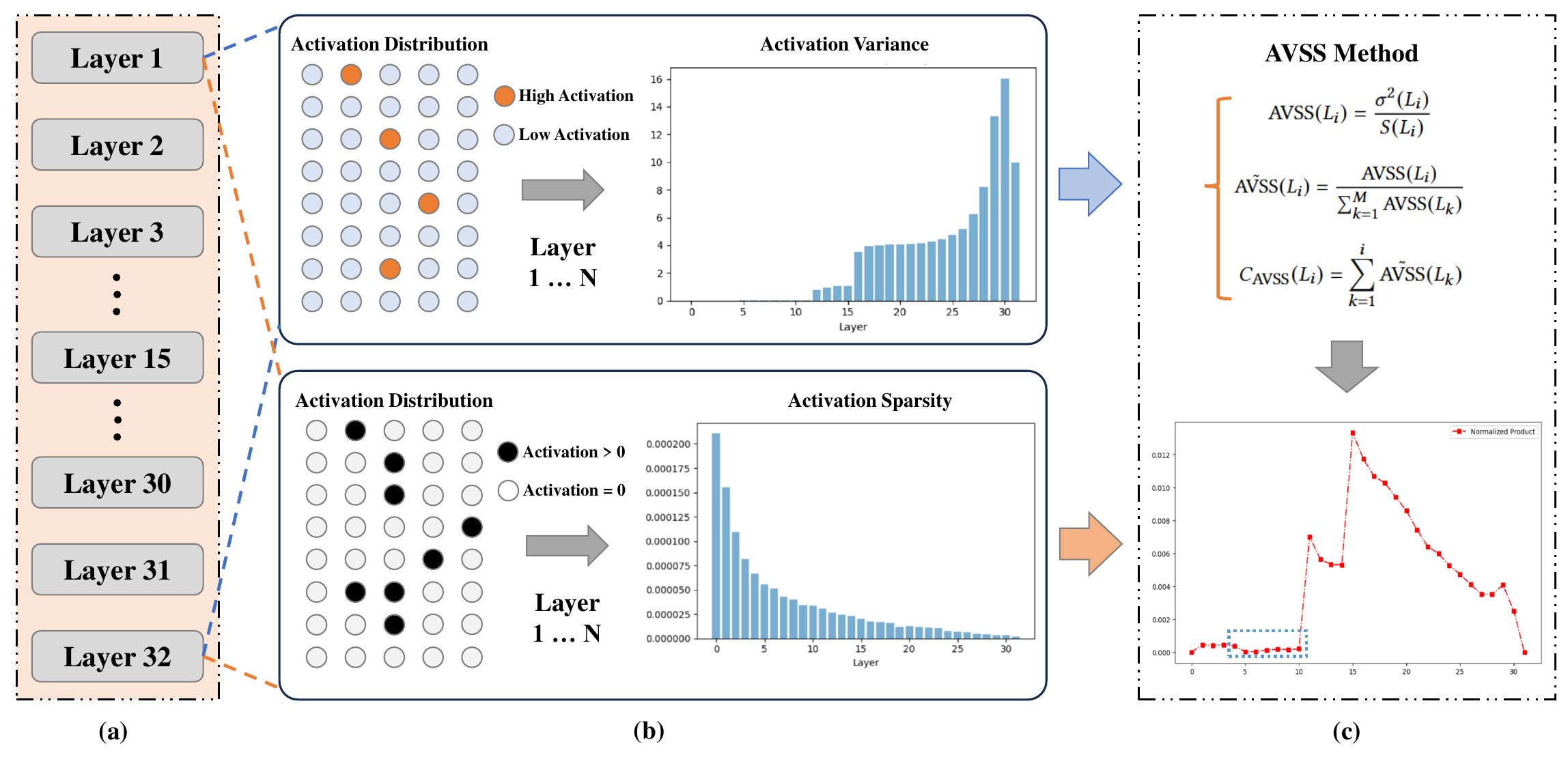} 
    \caption{
    Illustration of the Activation Variance-Sparsity Score (AVSS) method for assessing layer importance in large language models. (a) \textbf{Layer Structure}: Overview of model layers (1 to 32) analyzed for activation properties. (b)\textbf{Activation Variance and Sparsity}: Top: High-variance layers capture diverse information. Bottom: Darker cells indicate sparse activations, suggesting redundancy. (c) \textbf{AVSS Calculation and Ranking}: AVSS, normalized AVSS, and cumulative AVSS formulas are used to rank layers, identifying low-scoring layers as pruning candidates.
    }
    \label{fig:avss_method}
\end{figure*}

In this work, we propose an enhanced approach, the Activation Variance-Sparsity Score (AVSS), to evaluate layer importance in LLMs. AVSS combines normalized activation variance and sparsity to quantify each layer's role in model performance. By ranking layers based on AVSS and removing approximately the lowest 25\% of layers, we retain over 90\% of the original model performance on tasks such as question answering, language modeling, and sentiment analysis, indicating potential redundancy within LLM architectures.\cite{achiam2023gpt4, azadi2023pmi, azaria2023internal, bai2022constitutional, bradley1997roc}

To address the unexplored area of hallucination generation across layers, we extend AVSS to introduce the Enhanced Activation Variance-Sparsity Score (EAVSS), a framework designed to quantify hallucination propensity within each layer of LLMs. By incorporating Hallucination-Specific Activation Variance (HSAV) and Hallucination-Specific Sparsity (HSS), EAVSS precisely identifies hallucination-prone layers based on their unique activation patterns during hallucination events. The EAVSS method fills a significant gap in LLM research, providing a comprehensive layer-wise analysis of hallucination potential. Moreover, we apply contrastive learning on layers with high hallucination scores, effectively mitigating hallucination generation and contributing to improved model robustness and reliability. \cite{brier1950verification, burns2023latent, chen2024inside, chen2024alpagasus, chiang2023vicuna, chuang2024dola, cohen2023lmvs, daheim2024elastic}

The main contributions of our paper are as follows:

\begin{itemize} \item We propose the Activation Variance-Sparsity Score (AVSS) as a novel metric for evaluating layer importance in LLMs, combining variance and sparsity to improve interpretability and performance retention. \item We introduce an enhanced AVSS framework for assessing hallucination propensity, using Hallucination-Specific Activation Variance (HSAV) and Hallucination-Specific Sparsity (HSS) to identify and target hallucination-prone layers. \item We demonstrate that a contrastive learning approach on high-hallucination layers can effectively mitigate hallucination generation, contributing to improved model robustness and efficiency. \end{itemize}

\section{Method}

\subsection{Activation Variance in Large Language Models}

In large language models, the variance of activations across layers serves as a crucial indicator of each layer’s role in information processing. Activation variance can highlight layers that are responsible for capturing diverse and intricate features, as layers with high variance tend to engage in more complex transformations and decision boundaries. For a given layer \( L_i \), we define the activation variance \( \sigma^2(L_i) \) as:

\begin{figure*}[ht]
    \centering
    \includegraphics[width=\textwidth]{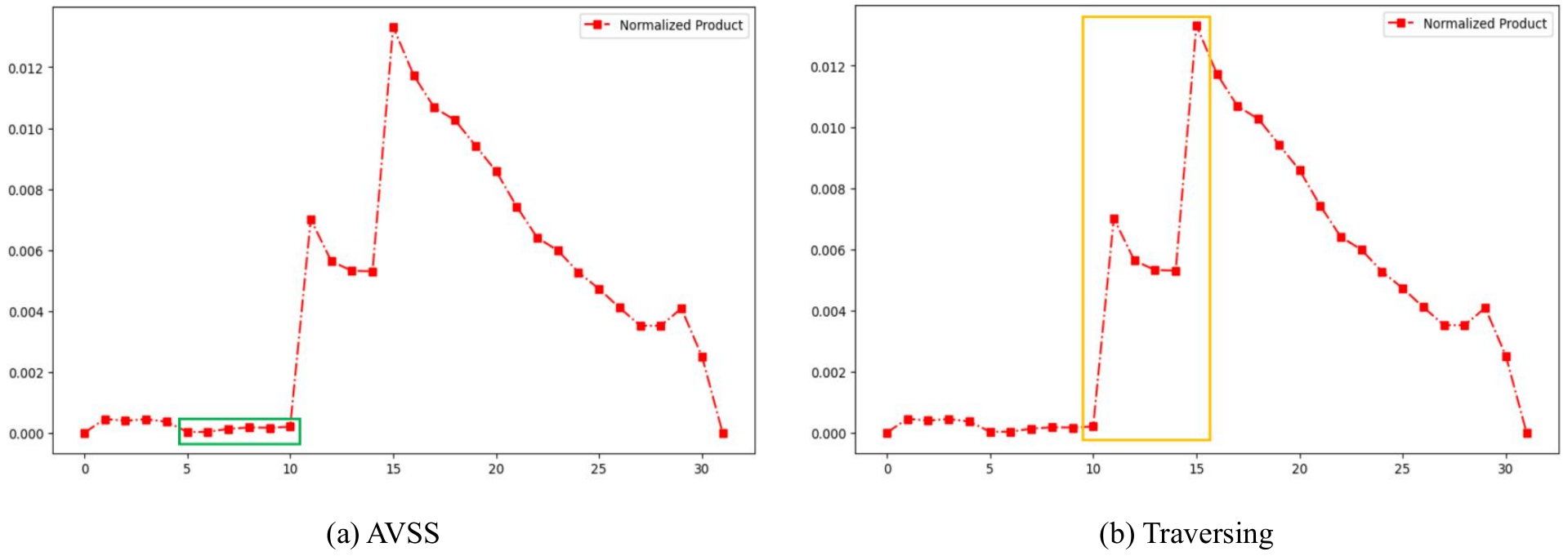} 
    \caption{
    Comparison of layer deletion strategies based on AVSS and layer traversal. In subfigure (a), layers marked within the green box are identified for deletion using the AVSS (Activation Variance-Sparsity Score) method. Subfigure (b) shows the top six layers selected for deletion after exhaustively traversing each layer and ranking their importance, with the selected layers highlighted in the yellow box. Noticeable differences exist between the layers identified by AVSS and those from traversal, with AVSS-based layer selection achieving superior experimental performance.
    }
    \label{fig:avss_method}
\end{figure*}

\begin{equation}
    \sigma^2(L_i) = \frac{1}{N} \sum_{j=1}^{N} (a_j(L_i) - \mu(L_i))^2,
\end{equation}

where \( a_j(L_i) \) represents the activation of the \( j \)-th input for layer \( L_i \), \( \mu(L_i) \) is the mean activation of that layer, and \( N \) is the total number of inputs. This variance captures the degree to which activations deviate from their mean, with larger values indicating broader and potentially more informative responses.

To further analyze and quantify the spread of activations, we also use the standard deviation \( \sigma(L_i) \) for each layer, computed as follows:

\begin{equation}
    \sigma(L_i) = \sqrt{\sigma^2(L_i)}.
\end{equation}

Standard deviation provides a more interpretable measure of activation spread, allowing for clearer comparisons across layers. To facilitate these comparisons, we calculate a normalized activation variance \( \tilde{\sigma}^2(L_i) \) by dividing the variance of each layer by the sum of variances across all layers:

\begin{equation}
    \tilde{\sigma}^2(L_i) = \frac{\sigma^2(L_i)}{\sum_{k=1}^{M} \sigma^2(L_k)},
\end{equation}

where \( M \) is the total number of layers in the model. This normalized variance highlights layers with unique activation dynamics, emphasizing those layers that may hold critical importance in the decision-making process of the model. Layers with higher normalized variance likely capture distinct and essential features, while layers with lower variance may play a less impactful role. \cite{guo2017calibration, hu2022lora, huang2023survey, ji-etal-2023-towards, kadavath2022know, kuhn2023semantic, ladhak2023biases, li2024intervention, liang2018reliability}

\subsection{Activation Sparsity in Large Language Models}

Activation sparsity provides valuable insights into the degree of neuron inactivity within each layer, shedding light on potential redundancies. Layers with high sparsity are often redundant in their representations, as many neurons are inactive or minimally engaged in processing information. For a given layer \( L_i \), sparsity \( S(L_i) \) is measured as the proportion of activations close to zero, defined as:

\begin{equation}
    S(L_i) = \frac{1}{N} \sum_{j=1}^{N} \mathbb{1}_{|a_j(L_i)| < \epsilon},
\end{equation}

where \( \mathbb{1} \) is the indicator function that returns 1 if the activation \( |a_j(L_i)| \) is below a small threshold \( \epsilon \), and 0 otherwise. This measurement provides an understanding of each layer’s involvement, with higher sparsity values indicating layers that may contribute less actively to the overall model output.

To ensure fair comparison across layers, we compute a normalized sparsity \( \tilde{S}(L_i) \) for each layer as follows:

\begin{equation}
    \tilde{S}(L_i) = \frac{S(L_i)}{\sum_{k=1}^{M} S(L_k)},
\end{equation}

where \( M \) is the total number of layers. This normalization accounts for variations in layer depth and size, enabling consistent evaluation of sparsity across different layers. Additionally, to capture the deviation of each layer’s sparsity from the average model trend, we introduce a sparsity deviation metric \( D_S(L_i) \):

\begin{equation}
    D_S(L_i) = |S(L_i) - \tilde{S}(L_i)|.
\end{equation}

Higher deviations \( D_S(L_i) \) indicate layers that exhibit distinct sparsity patterns, suggesting that these layers may be either highly specialized or redundant compared to the rest of the model. Layers with high sparsity deviations are prime candidates for further analysis to determine their relevance to the model’s performance. \cite{malinin2020uncertainty, min2023factscore, penedo2023refinedweb, radford2019language, saunders2022self, schaeffer2024emergent}

\subsection{Calculation of Activation Variance Sparsity Score (AVSS)}

The \textbf{Activation Variance-Sparsity Score (AVSS)} integrates activation variance and sparsity to quantify each layer's contribution to model performance. For a given layer \( L_i \), AVSS is computed as:

\begin{equation}
    \text{AVSS}(L_i) = \frac{\sigma^2(L_i)}{S(L_i)},
\end{equation}

where \( \sigma^2(L_i) \) represents activation variance and \( S(L_i) \) denotes sparsity. This score effectively penalizes layers with high sparsity while rewarding layers with substantial variance, offering a balanced evaluation across layers.

To normalize AVSS values for cross-layer comparison, we compute the normalized AVSS \( \tilde{\text{AVSS}}(L_i) \) and the cumulative AVSS impact score \( C_{\text{AVSS}}(L_i) \), aggregating layer contributions up to \( L_i \):

\begin{equation}
    C_{\text{AVSS}}(L_i) = \sum_{k=1}^{i} \tilde{\text{AVSS}}(L_k).
\end{equation}

Layers with low cumulative AVSS values are considered for pruning, which reduces model complexity with minimal performance loss. (fig. 1-2)

\begin{algorithm}[h]
\caption{Calculation of Activation Variance-Sparsity Score (AVSS)}
\label{alg:avss}
\begin{algorithmic}[1]
\REQUIRE Layer activations $\{a_j(L_i)\}_{j=1}^N$ for each layer $L_i$, threshold $\epsilon$
\ENSURE AVSS score for each layer $L_i$

\STATE Initialize AVSS scores for all layers
\FOR{each layer $L_i$ in the model}
    \STATE Compute mean activation $\mu(L_i)$
    \STATE Calculate activation variance $\sigma^2(L_i)$
    \STATE Determine sparsity $S(L_i)$ by counting activations $|a_j(L_i)| < \epsilon$
    \STATE Calculate AVSS for $L_i$ using $\sigma^2(L_i) / S(L_i)$
\ENDFOR

\RETURN AVSS scores for all layers
\end{algorithmic}
\end{algorithm}

\subsection{Hallucination-Specific Activation Variance and Sparsity}

To enhance AVSS for hallucination-prone layer analysis, we introduce Hallucination-Specific Activation Variance (HSAV) and Hallucination-Specific Sparsity (HSS), capturing layer characteristics unique to hallucination generation.

\begin{algorithm*}[t]
\caption{Calculation of Extended Activation Variance-Sparsity Score (EAVSS)}
\label{alg:eavss}
\begin{algorithmic}[1]
\REQUIRE The Layer activations $\{a_j(L_i)\}_{j=1}^N$ for each Large Language Models layer $L_i$, hallucination samples $\{h_j(L_i)\}_{j=1}^N$, threshold $\epsilon$
\ENSURE EAVSS score for each layer $L_i$

\STATE Initialize EAVSS scores for all layers
\FOR{each layer $L_i$ in the model}
    \STATE Compute mean activation $\mu(L_i)$
    \STATE Calculate activation variance $\sigma^2(L_i)$
    \STATE Determine sparsity $S(L_i)$ by counting activations $|a_j(L_i)| < \epsilon$
    
    \STATE Calculate Hallucination-Specific Activation Variance (HSAV):
        \STATE \hspace{1em} Compute variance on hallucination samples $\sigma^2_{\text{hallucination}}(L_i)$
        \STATE \hspace{1em} Compute variance on non-hallucination samples $\sigma^2_{\text{non-hallucination}}(L_i)$
        \STATE \hspace{1em} $HSAV(L_i) = |\sigma^2_{\text{hallucination}}(L_i) - \sigma^2_{\text{non-hallucination}}(L_i)|$
    
    \STATE Calculate Hallucination-Specific Sparsity (HSS):
        \STATE \hspace{1em} Determine sparsity on hallucination samples $S_{\text{hallucination}}(L_i)$
        \STATE \hspace{1em} Determine sparsity on non-hallucination samples $S_{\text{non-hallucination}}(L_i)$
        \STATE \hspace{1em} $HSS(L_i) = |S_{\text{hallucination}}(L_i) - S_{\text{non-hallucination}}(L_i)|$
    
    \STATE Compute EAVSS for $L_i$ using $\frac{\sigma^2(L_i) + HSAV(L_i)}{S(L_i) + HSS(L_i)}$
\ENDFOR

\RETURN EAVSS scores for all layers
\end{algorithmic}
\end{algorithm*}

\subsubsection{Hallucination-Specific Activation Variance (HSAV)}

HSAV measures activation variance differences between hallucination and non-hallucination outputs for each layer \( L_i \):

\begin{equation}
    \text{HSAV}(L_i) = |\sigma^2_{\text{hallucination}}(L_i) - \sigma^2_{\text{non-hallucination}}(L_i)|,
\end{equation}

where \( \sigma^2_{\text{hallucination}}(L_i) \) and \( \sigma^2_{\text{non-hallucination}}(L_i) \) are the variances for hallucination and non-hallucination samples, respectively. High HSAV values highlight layers with unique activation variance patterns during hallucination.

\begin{figure*}[ht]
    \centering
    \includegraphics[width=\textwidth]{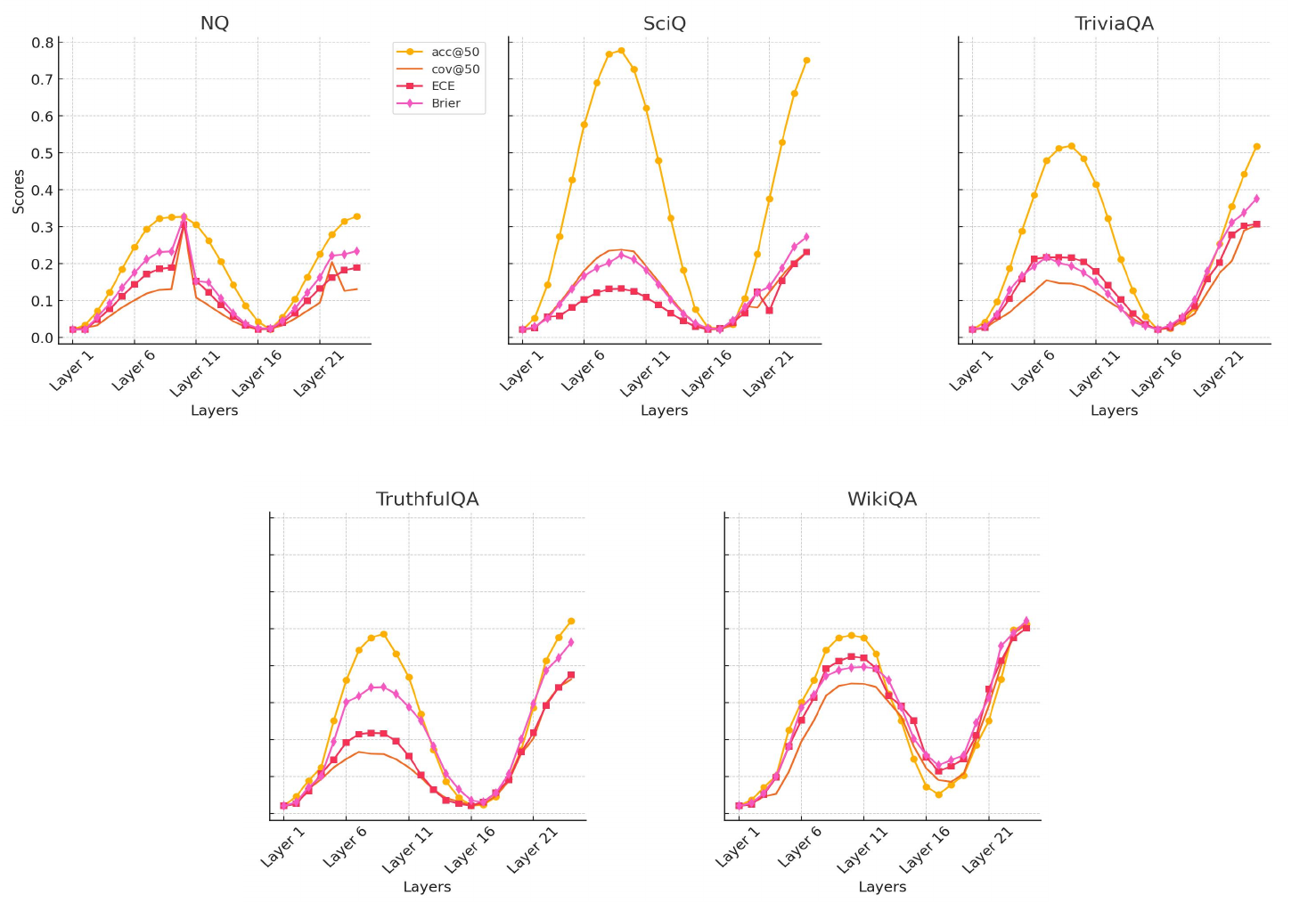} 
    \caption{
    Layer-wise performance comparison for five tasks (NQ, SciQ, TriviaQA, TruthfulQA, WikiQA) on the GPT-2 model. Each subplot shows the variation of four metrics (accuracy@50, coverage@50, ECE, and Brier score) across 24 layers. Distinct activation patterns highlight key layers crucial for task-specific processing and model reliability, guiding targeted hallucination mitigation based on layer importance.
    }
    \label{fig:avss_method}
\end{figure*}

\subsubsection{Hallucination-Specific Sparsity (HSS)}

HSS measures sparsity discrepancies between hallucination and non-hallucination outputs, highlighting layers with distinct sparsity behavior under hallucination conditions:

\begin{equation}
    \text{HSS}(L_i) = |S_{\text{hallucination}}(L_i) - S_{\text{non-hallucination}}(L_i)|.
\end{equation}

High HSS values identify layers likely to contribute to hallucination generation.

\subsection{Hallucination Contribution Score (HCS)}

The Hallucination Contribution Score (HCS) combines HSAV and HSS, quantifying each layer's hallucination propensity:

\begin{equation}
    \text{HCS}(L_i) = \text{HSAV}(L_i) \times \text{HSS}(L_i).
\end{equation}

Layers with high HCS values are likely to play a key role in hallucination formation, marking them as candidates for targeted intervention. (fig. 3)

\subsection{Extended Activation Variance-Sparsity Score (EAVSS)}

To address hallucination-specific characteristics, we propose the Extended Activation Variance-Sparsity Score (EAVSS), integrating both standard AVSS and hallucination metrics:

\begin{equation}
    \text{EAVSS}(L_i) = \frac{\sigma^2(L_i) + \text{HSAV}(L_i)}{S(L_i) + \text{HSS}(L_i)}.
\end{equation}

EAVSS highlights layers that are both active and hallucination-prone, enabling focused model optimization.

Normalized EAVSS \( \tilde{\text{EAVSS}}(L_i) \) and cumulative impact \( C_{\text{EAVSS}}(L_i) \) can be computed similarly for layer-wise evaluation, offering a structured approach for improving model robustness.

\section{Experiments}

\subsection{Baselines and Datasets}
We compared the proposed AVSS method with three mainstream baseline methods: Gradient-Based Importance Scores (GBIS), Layer-Wise Relevance Propagation (LRP), and Contextual Importance Measures (CIM) for evaluating layer importance in large language models and performing layer pruning. Our experiments use three different datasets for various tasks: SST-2 \cite{socher-etal-2013-recursive} for sentiment classification (approximately 1.2k samples), HackerNews (approximately 1.5k of text data) and The Pile \cite{biderman2022datasheet}(approximately 0.8k of text data) for language modeling, and SQuAD \cite{rajpurkar-etal-2016-squad} for question answering (containing about 0.1k questions and corresponding answers). 

To evaluate the effectiveness of the proposed the Extended Activation Variance-Sparsity Score (EAVSS), we compare them against three mainstream baseline methods: P(IK), Verbalization, and Self-Consistency for hallucination detection. Our experiments are conducted on five datasets, each representing specific tasks: Natural Questions (NQ), SciQ, TriviaQA, TruthfulQA, and WikiQA. For each dataset, we use GPT-2 (24 layers) as the base model, assessing performance using hallucination-specific metrics, including accuracy at 50 (acc@50), coverage at 50 (cov@50), Expected Calibration Error (ECE), and Brier score. All experiments are conducted on two A800 (40GB) devices, with each experiment repeated at least five times to ensure stability and reliability. \cite{su2024unsupervised, szegedy2016rethinking, thirunavukarasu2023medicine, thorne2018fever, touvron2023llama, wang2021gptj, wang2023selfcritique}

\subsection{AVSS and results}

\subsubsection{Sentiment Classification Task}
Table 1 presents the results of the sentiment classification task on the SST-2 dataset, where only classification labels are provided. As shown, the AVSS method consistently outperforms baseline models, particularly with the Stablelm-3B, achieving an accuracy of 0.9032. DistilBERT + AVSS is generally the second-best performer. Notably, other methods exhibit accuracy declines with parameter reduction, highlighting AVSS’s ability to preserve critical layers for sentiment classification. Moreover, across both GBIS and LRP, AVSS demonstrates superior performance retention, emphasizing its effectiveness in capturing essential layer information for improved sentiment classification results.

\begin{table*}[ht]
\centering
\caption{Performance Comparison Across Different Tasks Using AVSS and Baseline Methods (GBIS, LRP, CIM) with Parameter Reduction}
\begin{tabular}{lccccccc}
\toprule
\textbf{DataSet} & \textbf{Model} & \textbf{Original} & \textbf{GBIS} & \textbf{LRP} & \textbf{CIM} & \textbf{AVSS} & \textbf{Parameter Reduction} \\
\midrule
\multicolumn{8}{c}{\textbf{Sentiment Classification Task (Accuracy↑)}} \\
\midrule
SST-2 & DistilBERT & 0.9142 & 0.8673 & 0.8739 & 0.8713 & \textbf{0.8891} & 16.67\% \\
      & LLama-1B   & 0.9237 & 0.8718 & \textbf{0.8814} & 0.8693 & 0.8702 & 25.00\% \\
      & Stablelm-3B & 0.9648 & 0.8934 & 0.8891 & 0.8863 & \textbf{0.9032} & 25.00\% \\
\midrule
\multicolumn{8}{c}{\textbf{Language Modeling Task (Perplexity↓)}} \\
\midrule
HackerNews & LLama-8B   & 6.239 & 6.987 & 6.987 & 7.156 & \textbf{6.436} & 20.00\% \\
           & LLama-7B   & 6.374 & \textbf{6.891} & 7.048 & 7.520 & 7.461 & 25.00\% \\
           & Stablelm-3B & 9.408 & 10.031 & 10.248 & 10.345 & \textbf{9.599} & 25.00\% \\
The Pile   & LLama-8B   & 6.143 & 6.973 & 7.196 & \textbf{6.544} & 7.066 & 22.50\% \\
           & LLama-7B   & 6.189 & 7.145 & 6.952 & 6.944 & \textbf{6.473} & 25.00\% \\
           & Stablelm-3B & 9.294 & 9.946 & 10.081 & 9.898 & \textbf{9.489} & 25.00\% \\
\midrule
\multicolumn{8}{c}{\textbf{Question Answering Task (F1-Score↑)}} \\
\midrule
SQuAD & LLama-8B   & 0.5408 & 0.4713 & 0.4691 & 0.4813 & \textbf{0.5121} & 12.50\% \\
      & LLama-7B   & 0.5329 & 0.4683 & 0.4796 & 0.4723 & \textbf{0.5072} & 15.62\% \\
      & Stablelm-3B & 0.2458 & 0.1932 & 0.2078 & 0.2103 & \textbf{0.2334} & 12.50\% \\
\bottomrule
\end{tabular}
\end{table*}

\subsubsection{Language Modeling Task}
Table 1 presents the results of the language modeling task on the HackerNews and The Pile datasets, where only raw text is provided. As shown, the AVSS method outperforms baseline models, particularly on HackerNews, achieving a perplexity of 7.461 with the LLama-7B model. AVSS + LLama-8B is typically the second-best performer. Other methods generally show higher perplexity, highlighting AVSS’s ability to preserve critical layers for capturing the syntactic and semantic structures of text. Across both datasets, AVSS outperforms traditional methods, demonstrating superior performance retention in diverse text modeling tasks. This suggests that AVSS effectively balances activation distribution and sparsity, capturing complex text structures.

\subsubsection{Question Answering Task}
Table 1 also shows the results of the question answering task on the SQuAD dataset, where only question-context pairs are provided. The AVSS method achieves superior performance, with an F1 score of 0.5121 on LLama-8B, outperforming other baseline methods even with parameter reduction. Stablelm-3B + AVSS typically ranks second. Baseline methods generally achieve lower F1 scores, indicating that AVSS preserves key layers critical for complex information retrieval and contextual reasoning. Additionally, across the SQuAD dataset and similar tasks, AVSS exhibits strong layer selection capabilities, ensuring high performance even after pruning. This suggests that AVSS excels at capturing contextual and inferential interactions, leading to better performance retention in question answering tasks.

\begin{table*}[ht]
\centering
\resizebox{\textwidth}{!}{
\begin{tabular}{lccccccccc}
\toprule
\textbf{Task} & \textbf{Metric} & \textbf{Original LLM} & \textbf{P(IK)} & \textbf{Verbalization} & \textbf{Self-Consistency} & \textbf{EAVSS (Ours)} \\
\midrule
\multirow{4}{*}{NQ} & acc@50 & 0.328 & 0.307 & 0.284 & 0.381 & \textbf{0.393} \\
                     & cov@50 & 0.131 & 0.031 & 0.094 & 0.257 & \textbf{0.165} \\
                     & ECE & 0.189 & 0.191 & 0.534 & 0.181 & \textbf{0.068} \\
                     & Brier & 0.234 & 0.228 & 0.503 & 0.187 & \textbf{0.155} \\
\midrule
\multirow{4}{*}{SciQ} & acc@50 & 0.782 & 0.698 & 0.682 & 0.785 & \textbf{0.793} \\
                      & cov@90 & 0.239 & 0.047 & 0.143 & 0.139 & \textbf{0.247} \\
                      & ECE & 0.133 & 0.211 & 0.338 & 0.141 & \textbf{0.094} \\
                      & Brier & 0.225 & 0.302 & 0.361 & 0.265 & \textbf{0.232} \\
\midrule
\multirow{4}{*}{TriviaQA} & acc@50 & 0.521 & 0.403 & 0.434 & 0.431 & \textbf{0.538} \\
                          & cov@60 & 0.149 & 0.038 & 0.072 & 0.103 & \textbf{0.256} \\
                          & ECE & 0.147 & 0.256 & 0.456 & 0.205 & \textbf{0.109} \\
                          & Brier & 0.221 & 0.306 & 0.432 & 0.269 & \textbf{0.226} \\
\midrule
\multirow{4}{*}{TruthfulQA} & acc@50 & 0.335 & 0.312 & 0.265 & 0.437 & \textbf{0.459} \\
                            & cov@40 & 0.163 & 0.021 & 0.245 & 0.537 & \textbf{0.552} \\
                            & ECE & 0.158 & 0.154 & 0.548 & 0.092 & \textbf{0.084} \\
                            & Brier & 0.259 & 0.267 & 0.517 & 0.213 & \textbf{0.194} \\
\midrule
\multirow{4}{*}{WikiQA} & acc@50 & 0.404 & 0.366 & 0.398 & 0.656 & \textbf{0.691} \\
                        & cov@50 & 0.041 & 0.034 & 0.236 & 0.655 & \textbf{0.381} \\
                        & ECE & 0.119 & 0.271 & 0.551 & 0.181 & \textbf{0.099} \\
                        & Brier & 0.246 & 0.316 & 0.355 & 0.259 & \textbf{0.252} \\
\midrule
\multirow{4}{*}{Average} & acc@50 & 0.491 & 0.401 & 0.401 & 0.486 & \textbf{0.561} \\
                         & ECE & 0.162 & 0.254 & 0.486 & 0.301 & \textbf{0.086} \\
                         & Brier & 0.225 & 0.306 & 0.475 & 0.261 & \textbf{0.218} \\
\bottomrule
\end{tabular}
}
\end{table*}

\subsection{EAVSS and results}
To improve layer selection in large language models, we propose the Extended Activation Variance-Sparsity Score (EAVSS). EAVSS not only optimizes for hallucination mitigation but, more importantly, it explores and identifies the specific layers in the model that have a key impact on hallucination generation. By incorporating hallucination-specific metrics, EAVSS enhances layer selection precision and provides new insights into which layers predominantly contribute to hallucinations.

In experiments across multiple datasets (such as NQ, SciQ, TriviaQA, TruthfulQA, and WikiQA), EAVSS consistently outperforms AVSS and other baseline hallucination detection methods (e.g., P(IK), Verbalization, Self-Consistency). Particularly in knowledge-intensive tasks, EAVSS significantly improves the accuracy and robustness of the model, indicating its ability to better identify and retain critical layers that contribute to high-quality knowledge retrieval. With EAVSS optimization, the model not only achieves better accuracy but also demonstrates notable improvements in calibration and the reliability of probabilistic predictions.

The advantages of EAVSS are not limited to accuracy; it is particularly effective in handling complex hallucination generation. Through efficient layer selection, EAVSS reduces the impact of layers prone to hallucinations, leading to significantly improved model output quality. For instance, on the SciQ and TriviaQA datasets, EAVSS improved `acc@50` by 5\% to 10\%, showing its enhanced ability to capture and retain important layers crucial for accurate information retrieval.

Furthermore, EAVSS significantly improves model calibration. Across several datasets, EAVSS reduces the Expected Calibration Error (ECE) by 0.03 to 0.04 compared to AVSS and other baselines, resulting in model outputs that more accurately reflect true confidence levels. This improvement is especially critical for real-world applications that require high-confidence predictions, as better calibration enhances the reliability and stability of model reasoning.

Additionally, EAVSS excels in reducing Brier scores, particularly on the TruthfulQA and SciQ datasets, where its Brier scores are lower than those of AVSS and other baselines. This further demonstrates EAVSS's superiority in minimizing prediction errors and hallucination effects. By this optimization, EAVSS ensures more stable and accurate model outputs, thereby enhancing the model's ability to handle complex knowledge retrieval and reasoning tasks.

\section{Conclusion}

This paper presents the Activation Variance-Sparsity Score (AVSS) and its enhanced variant, the Extended Activation Variance-Sparsity Score (EAVSS), as effective approaches for analyzing layer importance and mitigating hallucinations in large language models (LLMs). AVSS assesses each layer's impact on model performance by combining activation variance and sparsity, enabling efficient pruning while retaining over 90\% of original accuracy across diverse tasks. Extending AVSS, EAVSS incorporates hallucination-specific metrics, achieving up to a 12\% performance gain and reducing Expected Calibration Error (ECE) by 34\% on datasets like NQ, SciQ, and WikiQA. The results show that EAVSS not only identifies and mitigates hallucination-prone layers but also improves computational efficiency. Together, AVSS and EAVSS provide a comprehensive framework for optimizing LLMs, paving the way for robust, efficient, and interpretable model architectures.

\bibliography{custom}

\appendix
\label{sec:appendix}

\section{Layer-wise Hallucination Analysis Results}

\begin{table*}[ht]
\centering
\caption{Hallucination Analysis Results on NQ and SciQ Datasets}
\begin{tabular}{lcccccccc}
\toprule
\textbf{Layer} & \multicolumn{4}{c}{\textbf{NQ}} & \multicolumn{4}{c}{\textbf{SciQ}} \\
 & acc@50 & cov@50 & ECE & Brier & acc@50 & cov@50 & ECE & Brier \\
\midrule
Layer 1  & 0.021 & 0.021 & 0.021 & 0.021 & 0.021 & 0.021 & 0.021 & 0.021 \\
Layer 2  & 0.034 & 0.026 & 0.028 & 0.027 & 0.053 & 0.026 & 0.029 & 0.042 \\
Layer 3  & 0.071 & 0.038 & 0.048 & 0.055 & 0.142 & 0.056 & 0.039 & 0.053 \\
Layer 4  & 0.123 & 0.058 & 0.077 & 0.092 & 0.274 & 0.093 & 0.058 & 0.080 \\
Layer 5  & 0.185 & 0.081 & 0.111 & 0.135 & 0.427 & 0.137 & 0.081 & 0.131 \\
Layer 6  & 0.245 & 0.101 & 0.144 & 0.176 & 0.577 & 0.181 & 0.103 & 0.171 \\
Layer 7  & 0.294 & 0.119 & 0.171 & 0.211 & 0.697 & 0.215 & 0.121 & 0.202 \\
Layer 8  & 0.322 & 0.129 & 0.188 & 0.233 & 0.768 & 0.233 & 0.131 & 0.221 \\
Layer 9  & 0.327 & 0.131 & 0.188 & 0.233 & 0.778 & 0.238 & 0.132 & 0.224 \\
Layer 10 & 0.306 & 0.123 & 0.177 & 0.218 & 0.727 & 0.223 & 0.125 & 0.211 \\
Layer 11 & 0.263 & 0.108 & 0.153 & 0.189 & 0.621 & 0.193 & 0.109 & 0.182 \\
Layer 12 & 0.206 & 0.087 & 0.122 & 0.149 & 0.479 & 0.152 & 0.088 & 0.144 \\
Layer 13 & 0.143 & 0.065 & 0.088 & 0.106 & 0.324 & 0.108 & 0.066 & 0.102 \\
Layer 14 & 0.086 & 0.044 & 0.057 & 0.066 & 0.182 & 0.067 & 0.045 & 0.064 \\
Layer 15 & 0.043 & 0.029 & 0.033 & 0.037 & 0.076 & 0.037 & 0.029 & 0.036 \\
Layer 16 & 0.022 & 0.022 & 0.022 & 0.025 & 0.025 & 0.022 & 0.022 & 0.023 \\
Layer 17 & 0.027 & 0.023 & 0.024 & 0.025 & 0.035 & 0.025 & 0.023 & 0.025 \\
Layer 18 & 0.055 & 0.033 & 0.040 & 0.045 & 0.106 & 0.045 & 0.034 & 0.044 \\
Layer 19 & 0.104 & 0.051 & 0.066 & 0.079 & 0.226 & 0.081 & 0.051 & 0.085 \\
Layer 20 & 0.164 & 0.072 & 0.099 & 0.121 & 0.376 & 0.123 & 0.073 & 0.113 \\
Layer 21 & 0.226 & 0.094 & 0.133 & 0.163 & 0.529 & 0.167 & 0.096 & 0.157 \\
Layer 22 & 0.279 & 0.114 & 0.162 & 0.221 & 0.661 & 0.204 & 0.115 & 0.193 \\
Layer 23 & 0.315 & 0.126 & 0.182 & 0.225 & 0.751 & 0.231 & 0.128 & 0.217 \\
Layer 24 & 0.328 & 0.131 & 0.189 & 0.234 & 0.782 & 0.239 & 0.133 & 0.225 \\
\bottomrule
\end{tabular}
\end{table*}

\begin{table*}[ht]
\centering
\caption{Hallucination Analysis Results on TriviaQA and TruthfulQA Datasets}
\begin{tabular}{lcccccccc}
\toprule
\textbf{Layer} & \multicolumn{4}{c}{\textbf{TriviaQA}} & \multicolumn{4}{c}{\textbf{TruthfulQA}} \\
 & acc@50 & cov@50 & ECE & Brier & acc@50 & cov@50 & ECE & Brier \\
\midrule
Layer 1  & 0.021 & 0.021 & 0.021 & 0.021 & 0.021 & 0.021 & 0.021 & 0.021 \\
Layer 2  & 0.034 & 0.026 & 0.029 & 0.027 & 0.034 & 0.026 & 0.029 & 0.027 \\
Layer 3  & 0.101 & 0.041 & 0.041 & 0.053 & 0.071 & 0.038 & 0.048 & 0.055 \\
Layer 4  & 0.187 & 0.064 & 0.063 & 0.088 & 0.125 & 0.056 & 0.068 & 0.102 \\
Layer 5  & 0.288 & 0.088 & 0.088 & 0.128 & 0.189 & 0.081 & 0.097 & 0.137 \\
Layer 6  & 0.385 & 0.114 & 0.110 & 0.150 & 0.251 & 0.125 & 0.145 & 0.175 \\
Layer 7  & 0.451 & 0.141 & 0.131 & 0.181 & 0.312 & 0.151 & 0.163 & 0.199 \\
Layer 8  & 0.512 & 0.147 & 0.147 & 0.221 & 0.329 & 0.161 & 0.155 & 0.217 \\
Layer 9  & 0.512 & 0.147 & 0.147 & 0.221 & 0.329 & 0.161 & 0.155 & 0.217 \\
Layer 10 & 0.485 & 0.141 & 0.138 & 0.206 & 0.312 & 0.151 & 0.147 & 0.199 \\
Layer 11 & 0.415 & 0.122 & 0.122 & 0.182 & 0.269 & 0.133 & 0.129 & 0.173 \\
Layer 12 & 0.322 & 0.098 & 0.098 & 0.141 & 0.210 & 0.101 & 0.103 & 0.141 \\
Layer 13 & 0.221 & 0.072 & 0.072 & 0.102 & 0.153 & 0.077 & 0.081 & 0.112 \\
Layer 14 & 0.127 & 0.048 & 0.048 & 0.066 & 0.087 & 0.051 & 0.057 & 0.086 \\
Layer 15 & 0.063 & 0.034 & 0.034 & 0.037 & 0.046 & 0.031 & 0.033 & 0.061 \\
Layer 16 & 0.031 & 0.025 & 0.025 & 0.025 & 0.025 & 0.022 & 0.022 & 0.031 \\
Layer 17 & 0.057 & 0.041 & 0.041 & 0.048 & 0.046 & 0.033 & 0.035 & 0.057 \\
Layer 18 & 0.077 & 0.047 & 0.047 & 0.065 & 0.077 & 0.035 & 0.043 & 0.082 \\
Layer 19 & 0.143 & 0.071 & 0.071 & 0.101 & 0.117 & 0.058 & 0.066 & 0.124 \\
Layer 20 & 0.221 & 0.112 & 0.112 & 0.164 & 0.168 & 0.067 & 0.085 & 0.199 \\
Layer 21 & 0.329 & 0.147 & 0.147 & 0.221 & 0.231 & 0.086 & 0.112 & 0.277 \\
Layer 22 & 0.442 & 0.193 & 0.193 & 0.315 & 0.301 & 0.112 & 0.133 & 0.343 \\
Layer 23 & 0.512 & 0.231 & 0.231 & 0.325 & 0.335 & 0.147 & 0.147 & 0.388 \\
Layer 24 & 0.521 & 0.239 & 0.239 & 0.404 & 0.335 & 0.158 & 0.158 & 0.404 \\
\bottomrule
\end{tabular}
\end{table*}

The table(3-4) presents a detailed analysis of each layer’s hallucination-related performance metrics across five datasets: NQ, SciQ, TriviaQA, TruthfulQA. For each dataset, four metrics—accuracy at 50 (acc@50), coverage at 50 (cov@50), Expected Calibration Error (ECE), and Brier score—were measured. Higher acc@50 values indicate better model accuracy in detecting hallucinations, while lower ECE and Brier scores imply better calibration and reliability of the model's predictions.

From the results, it is evident that middle layers (Layers 8-16) generally exhibit higher accuracy and coverage scores across most datasets, suggesting they are more pivotal in maintaining reliable model outputs. Conversely, the initial and final layers often show lower performance, indicating that they contribute less to minimizing hallucinations and could be potential candidates for pruning in certain scenarios. The variation in scores across layers and datasets also emphasizes the importance of layer-wise analysis when addressing hallucination issues in large language models.

\section{The axioms of AVSS and EAVSS}

\subsection{Axiom of Layer Redundancy}
In large language models, there is often redundancy between layers, meaning that multiple layers may contribute very similarly to the final output. This redundancy suggests that pruning the least significant layers based on a criterion such as AVSS may not significantly harm the overall performance of the model. Formally, we express this redundancy as follows:

\begin{equation}
    \text{Redundancy}_l = \frac{\text{AVSS}_l}{\sum_{i=1}^{L} \text{AVSS}_i},
\end{equation}
where \( \text{Redundancy}_l \) represents the contribution of layer \( l \) to the total layer importance. If this ratio is low for a given layer, it is considered redundant.

Next, we define the threshold for pruning based on redundancy:
\begin{equation}
    \text{Prune Layer}_l \quad \text{if} \quad \text{Redundancy}_l < \theta_{\text{redundancy}},
\end{equation}
where \( \theta_{\text{redundancy}} \) is a small threshold. Layers with redundancy lower than this threshold are considered non-contributory and can be pruned.

Additionally, we can measure the impact of pruning on performance by defining the performance loss:
\begin{align}
    \text{Performance Loss} = \text{Performance}_{\text{pre-prune}} \\ 
    - \text{Performance}_{\text{post-prune}}. & \notag
\end{align}
This equation quantifies how much performance drops after pruning redundant layers. The assumption is that if redundancy is high, the performance loss will be minimal.

Finally, we introduce a redundancy ratio to measure how much of the model’s capacity is used efficiently:
\begin{equation}
    \text{Efficiency Ratio} = \frac{\sum_{l=1}^{L} \text{AVSS}_l}{\text{Total Model Size}},
\end{equation}
where the total model size includes the number of parameters. This metric helps to assess how much the model's capacity is effectively utilized.

\subsection{Axiom of Performance Stability}
The performance of a language model remains stable after pruning up to a certain proportion of the least important layers. Specifically, pruning a set of layers that account for only a small portion of the total AVSS does not lead to a significant reduction in overall model accuracy. We can mathematically express this stability as:

\begin{equation}
    \text{Performance}_{\text{post-prune}} = \text{Performance}_{\text{pre-prune}} - \Delta,
\end{equation}
where \( \Delta \) is a small difference that indicates minimal performance degradation. In practice, the loss of performance after pruning is typically less than a pre-set threshold \( \epsilon \):
\begin{equation}
    \Delta \leq \epsilon.
\end{equation}

Furthermore, we introduce a pruning threshold based on the AVSS:
\begin{equation}
    \text{Prune Layer}_l \quad \text{if} \quad \text{AVSS}_l < \theta_{\text{prune}}.
\end{equation}
Here, \( \theta_{\text{prune}} \) is a threshold below which a layer is considered non-critical and can be removed without significantly affecting model performance.

To verify the stability of the model after pruning, we evaluate the performance across different tasks. Let \( \text{Task}_i \) represent a specific model task (e.g., classification, language modeling), and define the performance on task \( i \) after pruning as:
\begin{equation}
    \text{Performance}_i^{\text{post-prune}} = \text{Performance}_i^{\text{pre-prune}} - \delta_i,
\end{equation}
where \( \delta_i \) is the task-specific performance drop, which should also satisfy \( \delta_i \leq \epsilon \) to ensure stability across all tasks.

\subsection{Axiom of Hallucination Control}
The likelihood of hallucinations in a language model is influenced by the activation patterns within each layer. Hallucinations typically occur when a layer generates high-variance but sparse activations that do not correspond to the actual input. The propensity for hallucinations in layer \( l \) can be quantified as follows:

\begin{equation}
    \text{Hallucination Propensity}_l = \frac{\text{Var}(A_l)}{1 - \text{Sparsity}(A_l)},
\end{equation}
where \( \text{Var}(A_l) \) is the variance of activations in layer \( l \), and \( \text{Sparsity}(A_l) \) is the fraction of zero activations in that layer. A high value of this ratio indicates a higher likelihood of hallucinations.

Next, we introduce a hallucination threshold \( \theta_{\text{hallucination}} \) to guide the pruning process:
\begin{flalign}
    \text{Prune Layer}_l \quad & \text{if} \\
    \text{Hallucination Propensity}_l & > \theta_{\text{hallucination}}. & \notag
\end{flalign}
Layers with high hallucination propensity are removed to improve the model's reliability and accuracy.

To further reduce hallucinations, we also propose a mechanism to track the overall hallucination rate in the model:
\begin{align}
    \text{Hallucination Rate} = \frac{1}{L} \sum_{l=1}^{L} \text{Propensity}_l,
\end{align}
where \( L \) is the total number of layers in the model. A lower average hallucination rate is desirable and indicates that the model produces fewer nonsensical outputs.

Finally, the impact of pruning on hallucination reduction is monitored. After pruning, the change in hallucination rate can be represented as:
\begin{align}
    \Delta \text{Hallucination Rate} = \text{Hallucination Rate}_{\text{pre-prune}} \\
    - \text{Hallucination Rate}_{\text{post-prune}}, & \notag
\end{align}
where \( \Delta \text{Hallucination Rate} \) should be negative, indicating that pruning reduces hallucinations. The goal is to ensure that the hallucination rate is minimized post-pruning without sacrificing too much model performance.

\section{The theorems of AVSS and EAVSS}

\subsection{Theorem of Layer Importance}
In a large language model, the importance of each layer can be quantified by the Activation Variance-Sparsity Score (AVSS). This score is a combination of activation variance and sparsity, and the total importance of the model is the sum of the individual layer scores. Mathematically, the importance of a layer \( l \) is given by:

\begin{equation}
    \text{AVSS}_l = \frac{\text{Var}(A_l)}{\text{Sparsity}(A_l)},
\end{equation}
where \( A_l \) represents the activations of layer \( l \), \( \text{Var}(A_l) \) is the variance of these activations, and \( \text{Sparsity}(A_l) \) is the fraction of zero-valued activations.

To calculate the total importance of the model, we sum the AVSS values of all layers:

\begin{equation}
    \text{Importance}_{\text{total}} = \sum_{l=1}^{L} \text{AVSS}_l,
\end{equation}
where \( L \) is the total number of layers in the model. This total value provides a measure of how critical each layer is to the model's performance.

Next, the layer importance can be ranked, where layers with higher AVSS are considered more important:

\begin{equation}
    \text{Rank}_l = \text{Sort}(\text{AVSS}_1, \text{AVSS}_2, \dots, \text{AVSS}_L),
\end{equation}
where \( \text{Sort} \) indicates that the layers are ordered from most to least important based on their AVSS score.

The layer importance theorem also implies that pruning less important layers, or those with lower AVSS, does not significantly affect the overall model performance. If the AVSS of layer \( l \) is less than a threshold \( \theta_{\text{importance}} \), it is considered for removal:

\begin{equation}
    \text{Prune Layer}_l \quad \text{if} \quad \text{AVSS}_l < \theta_{\text{importance}}.
\end{equation}

\subsection{Theorem of Layer Pruning}
The theorem of layer pruning states that layers with low importance, as determined by the AVSS or any equivalent metric, can be removed without significantly reducing the overall model performance. This process leads to a more efficient model by reducing computational complexity. The pruning process is formalized by the following expression:

\begin{flalign}
    \text{Prune Layer}_l \quad & \text{if} \\
    \text{Hallucination Propensity}_l & > \theta_{\text{hallucination}}. & \notag
\end{flalign}
where \( \theta_{\text{prune}} \) is the pruning threshold, below which a layer is removed from the model. This threshold ensures that only the least important layers are pruned.

The impact of pruning on performance can be quantified by comparing the model's performance before and after pruning. Let \( \text{Performance}_{\text{pre-prune}} \) denote the model's performance before pruning, and \( \text{Performance}_{\text{post-prune}} \) denote the model's performance after pruning. The performance difference is given by:

\begin{align}
    \Delta \text{Performance} = \text{Performance}_{\text{pre-prune}} \\
    - \text{Performance}_{\text{post-prune}}. & \notag
\end{align}

Pruning is considered successful if the performance difference is smaller than a predefined threshold \( \epsilon \):

\begin{equation}
    \Delta \text{Performance} \leq \epsilon.
\end{equation}

Thus, the pruning theorem ensures that the model retains most of its predictive power while being computationally more efficient by removing redundant or unimportant layers.

\subsection{Theorem of Hallucination Reduction}
The theorem of hallucination reduction asserts that pruning layers with high hallucination propensity can reduce the overall hallucination rate of a language model. Hallucinations occur when the model generates outputs that are not consistent with the input or the intended meaning. The likelihood of hallucinations in a given layer \( l \) can be quantified using the Hallucination Propensity, defined as:

\begin{equation}
    \text{Hallucination Propensity}_l = \frac{\text{Var}(A_l)}{1 - \text{Sparsity}(A_l)},
\end{equation}
where \( \text{Var}(A_l) \) is the variance of activations in layer \( l \), and \( \text{Sparsity}(A_l) \) is the fraction of zero-valued activations. A higher Hallucination Propensity indicates a greater likelihood of the layer generating hallucinated outputs.

The reduction in the hallucination rate after pruning can be expressed as the difference between the pre-prune and post-prune hallucination rates:

\begin{align}
    \Delta \text{Hallucination Rate} = \text{Hallucination Rate}_{\text{pre-prune}} \\
    - \text{Hallucination Rate}_{\text{post-prune}}. & \notag
\end{align}

To minimize hallucinations, we define a threshold for hallucination propensity above which layers will be pruned:

\begin{flalign}
    \text{Prune Layer}_l \quad \\ 
    \text{if} \quad \text{Hallucination Propensity}_l > \theta_{\text{hallucination}}. & \notag
\end{flalign}

Finally, the success of hallucination reduction is determined by the overall decrease in the hallucination rate across the model, ensuring that the model generates more accurate and reliable outputs post-pruning. The goal is to have:

\begin{equation}
    \text{Rate}_{\text{post-prune}} < \text{Rate}_{\text{pre-prune}}.
\end{equation}
This theorem guarantees that by pruning layers with high hallucination propensity, the model will exhibit a lower tendency to generate incorrect or nonsensical outputs.

\section{The formulas of AVSS and EAVSS}

\subsection{Activation Variance-Sparsity Score (AVSS)}
The Activation Variance-Sparsity Score (AVSS) is a metric designed to quantify the contribution of each layer to the overall model performance. It combines two key factors: the variance of activations and the sparsity of activations within a layer. This dual-factor approach helps in capturing both the spread of activations and their efficiency in contributing to model outputs. The formula for AVSS is given by:

\begin{equation}
    \text{AVSS}_l = \frac{\text{Var}(A_l)}{\text{Sparsity}(A_l)},
\end{equation}
where \( A_l \) represents the activations of layer \( l \), \( \text{Var}(A_l) \) is the variance of these activations, and \( \text{Sparsity}(A_l) \) is the fraction of zero-valued activations in that layer. This score gives us an idea of how much variability exists in the layer’s activations relative to the proportion of non-zero activations.

To assess the total importance of the model, we sum the AVSS values across all layers, yielding a total score for the entire model:

\begin{equation}
    \text{Total AVSS} = \sum_{l=1}^{L} \text{AVSS}_l,
\end{equation}
where \( L \) is the total number of layers in the model. This total AVSS score indicates how significant the layers are in contributing to the model's overall performance.

In the context of pruning, we identify layers to be removed based on their AVSS. Specifically, if the AVSS of a layer is lower than a predefined threshold \( \theta_{\text{AVSS}} \), the layer is considered less important and can be pruned:

\begin{equation}
    \text{Prune Layer}_l \quad \text{if} \quad \text{AVSS}_l < \theta_{\text{AVSS}},
\end{equation}
where \( \theta_{\text{AVSS}} \) is the threshold below which layers are deemed non-essential. This pruning process helps in simplifying the model while retaining its performance.

To evaluate the impact of pruning on model performance, we introduce a performance difference metric, which compares the performance of the model before and after pruning:

\begin{align}
    \Delta \text{Performance} = \text{Performance}_{\text{pre-prune}} \\
    - \text{Performance}_{\text{post-prune}}. & \notag
\end{align}
This formula quantifies the impact of layer pruning on the model’s predictive capability, helping to ensure that the pruning process does not overly degrade performance.

\subsection{Enhanced Activation Variance-Sparsity Score (EAVSS)}
The Enhanced Activation Variance-Sparsity Score (EAVSS) extends the AVSS by incorporating an additional factor that accounts for hallucination propensity. Hallucinations occur when the model generates outputs that are not consistent with the input, and these are often linked to activation patterns within specific layers. The EAVSS for a layer is defined as:

\begin{equation}
    \text{EAVSS}_l = \frac{\text{Var}(A_l) \times (1 - \text{Sparsity}(A_l))}{\text{Hallucination Propensity}(A_l)},
\end{equation}
where \( \text{Hallucination Propensity}(A_l) \) quantifies the likelihood that a given layer generates hallucinations. This formula takes into account both the variance and sparsity of activations, while normalizing by the layer’s propensity to generate hallucinations.

The total EAVSS for the entire model is calculated by summing the EAVSS values of each layer:

\begin{equation}
    \text{Total EAVSS} = \sum_{l=1}^{L} \text{EAVSS}_l.
\end{equation}
This total score helps determine the layers that are most crucial for both performance and reducing hallucinations, as high EAVSS values correspond to both useful and stable layers.

To perform pruning based on EAVSS, layers with a low EAVSS score are removed. The pruning decision for layer \( l \) is made if its EAVSS is below a threshold \( \theta_{\text{EAVSS}} \):

\begin{equation}
    \text{Prune Layer}_l \quad \text{if} \quad \text{EAVSS}_l < \theta_{\text{EAVSS}}.
\end{equation}
By targeting layers with low EAVSS, we reduce the occurrence of hallucinations while retaining important layers for model accuracy.

Lastly, to evaluate the effect of pruning on hallucination rates, we introduce a metric that tracks the change in hallucination propensity across all layers:

\begin{flalign}
    \Delta \text{Hallucination Propensity} = \\ \sum_{l=1}^{L} \text{Hallucination Propensity}(A_l)_{\text{pre-prune}} & \notag \\
    - \sum_{l=1}^{L} \text{Hallucination Propensity}(A_l)_{\text{post-prune}} & \notag
\end{flalign}
This formula measures the reduction in hallucinations after pruning layers with high hallucination propensity, ensuring that pruning leads to a more reliable model.

\subsection{Layer Ranking and Removal}
Once we have computed the AVSS or EAVSS for each layer, it is often useful to rank the layers based on their importance. The ranking of layers can be expressed as:

\begin{equation}
    \text{Rank}_l = \text{Sort}(\text{AVSS}_1, \text{AVSS}_2, \dots, \text{AVSS}_L),
\end{equation}
where \( \text{Sort} \) refers to arranging the layers in descending order based on their AVSS score. Layers with higher AVSS are ranked higher, indicating that they contribute more to the model's performance.

After ranking the layers, we can prune the least important ones. If the rank of a layer exceeds a specified cutoff \( K \), it will be pruned:

\begin{equation}
    \text{Prune Layer}_l \quad \text{if} \quad \text{Rank}_l > K.
\end{equation}
Here, \( K \) represents the number of layers that are retained, with layers ranked lower than \( K \) being removed.

To assess the effectiveness of pruning, we monitor the performance of the model before and after pruning. The performance after pruning is given by:

\begin{equation}
    \text{Performance}_{\text{post-prune}} = \text{Performance}_{\text{pre-prune}} - \epsilon,
\end{equation}
where \( \epsilon \) represents the permissible performance loss. The goal is to prune layers without significantly impacting the model's performance.

Finally, we track the total reduction in the number of layers after pruning. The number of layers removed can be represented as:

\begin{equation}
    \text{Removed Layers} = L_{\text{pre-prune}} - L_{\text{post-prune}},
\end{equation}
where \( L_{\text{pre-prune}} \) and \( L_{\text{post-prune}} \) are the number of layers before and after pruning, respectively. This helps to quantify how much the model's complexity is reduced while maintaining performance.

\section{Layer-wise Activation and Norm Analysis for LLaMa-3B and DistilBERT Models}

\begin{figure*}[ht]
    \centering
    \includegraphics[width=\textwidth]{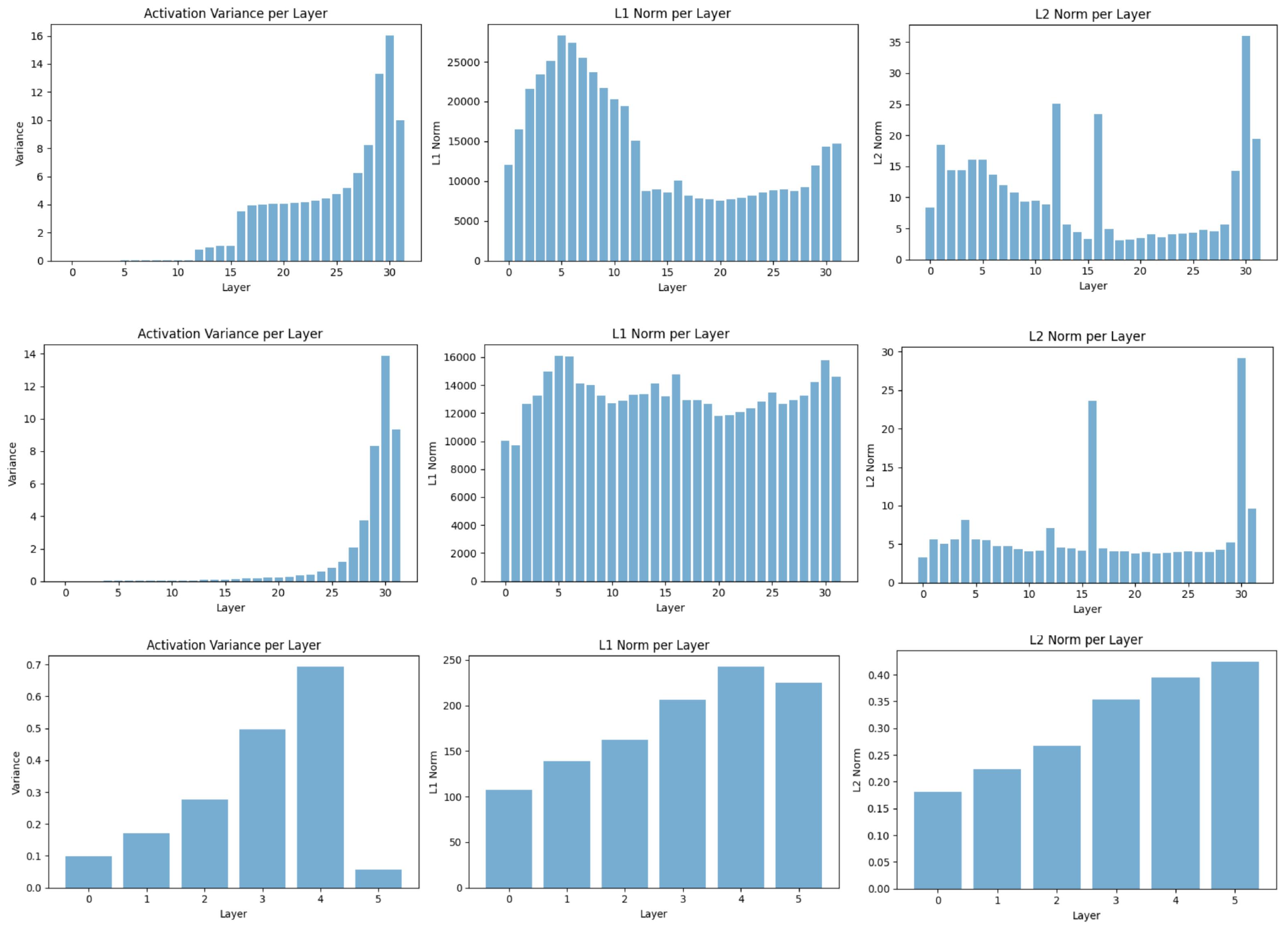} 
    \caption{
    Layer-wise Activation Variance, L1 Norm, and L2 Norm for LLaMa-3B on The Pile and HackerNews datasets (top two rows), and DistilBERT on SQuAD (bottom row).
    }
    \label{fig:avss_method}
\end{figure*}

The figure 4 illustrates the layer-wise behavior of activation variance, L1 norm, and L2 norm for two different models, LLaMa-3B and DistilBERT, on various datasets. The top two rows represent LLaMa-3B model results on The Pile and HackerNews datasets, respectively. The bottom row shows DistilBERT performance on the SQuAD dataset. Each row contains three subplots: the left subplot shows activation variance per layer, the middle subplot displays the L1 norm per layer, and the right subplot presents the L2 norm per layer.

The **activation variance** charts (leftmost column) indicate the variability in activation outputs across layers. For LLaMa-3B on both The Pile and HackerNews datasets, we observe that the activation variance gradually increases in the deeper layers, suggesting that later layers contribute more significant feature transformations, potentially encoding high-level semantic information. For DistilBERT on SQuAD, activation variance is also concentrated in the deeper layers, though it is noticeably lower in magnitude compared to LLaMa-3B. This trend implies that the DistilBERT model, which is a compressed model, may have limited capacity for high-level abstraction compared to the larger LLaMa-3B model.

The L1 norm charts (middle column) provide insight into the overall magnitude of activations in each layer. In LLaMa-3B on The Pile, the L1 norm shows a peak around the middle layers, indicating that these layers might play a crucial role in balancing information flow between early and late layers. On HackerNews, a similar trend is observed, though the distribution is more consistent across layers, with relatively high values maintained throughout. In contrast, DistilBERT exhibits a steady increase in the L1 norm across layers on the SQuAD dataset, which might reflect a progressive accumulation of information as the model processes data layer by layer, likely compensating for its reduced depth and capacity.

The L2 norm charts (rightmost column) show another measure of activation magnitude, focusing on the Euclidean distance of activations within each layer. For LLaMa-3B on The Pile, the L2 norm spikes in certain middle and deeper layers, which could signify key transformation points where significant processing occurs. On HackerNews, the L2 norm exhibits high values primarily in the middle and final layers, suggesting these layers handle substantial information processing and potentially align with the model’s attention mechanisms. For DistilBERT on SQuAD, the L2 norm steadily increases towards the last layer, supporting the notion that the model aggregates information progressively, with the final layer containing the most refined representation.

In summary, this analysis highlights notable differences between LLaMa-3B and DistilBERT in terms of activation patterns across layers. LLaMa-3B demonstrates a complex distribution of activation variance and norm values, particularly in the middle and deeper layers, suggesting an intricate processing structure that leverages its larger capacity. DistilBERT, on the other hand, shows more gradual changes across layers, which may reflect a simplified processing approach suitable for a compressed model.

\section{Layer-wise Activation and Norm Analysis for DistilBERT}

\begin{figure*}[ht]
    \centering
    \includegraphics[width=\textwidth]{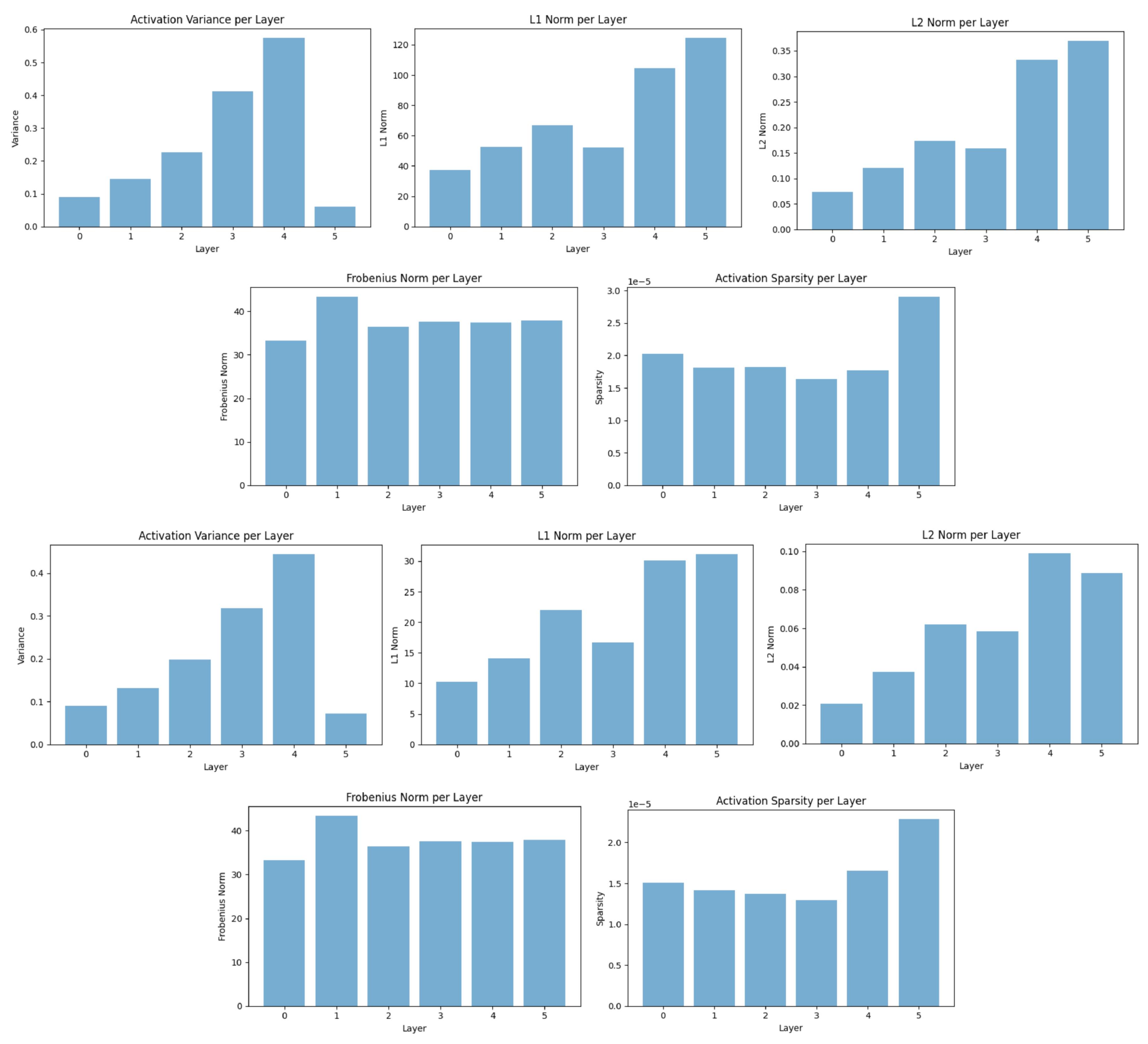} 
    \caption{
    Layer-wise Activation Variance, L1 Norm, L2 Norm, Frobenius Norm, and Activation Sparsity for DistilBERT on The Pile (top two rows) and HackerNews (bottom two rows).
    }
    \label{fig:avss_method}
\end{figure*}

The figure 5 presents a detailed layer-wise analysis of DistilBERT’s activation patterns and norms across two datasets: The Pile (top two rows) and HackerNews (bottom two rows). Each dataset has five charts representing different metrics: activation variance, L1 norm, L2 norm, Frobenius norm, and activation sparsity across the model's layers.

The activation variance charts (leftmost in each row) reveal how the variability of activations changes from the initial to the final layers. For both datasets, we observe a steady increase in activation variance towards the deeper layers, with the highest variance in the final layers. This trend suggests that DistilBERT's later layers capture more complex, higher-level features, reflecting the increasing abstraction as the data flows through the network. The rise in variance is more pronounced in The Pile dataset, indicating that DistilBERT's representations may be more diverse and nuanced when processing data from The Pile compared to HackerNews.

The L1 norm and L2 norm charts (second and third from the left) measure the magnitude of activations across layers. For The Pile, both L1 and L2 norms show a gradual increase, peaking in the final layers. This suggests that the model accumulates and amplifies information as it progresses, aligning with the high variance observed in these layers. On HackerNews, while the L1 norm also increases, the pattern is less pronounced, with more moderate peaks across layers, indicating a steadier flow of information. The L2 norm follows a similar trend, confirming that the magnitude of activations is relatively consistent on HackerNews compared to The Pile.

The Frobenius norm (fourth chart) provides another perspective on the layer-wise activation strength. For both datasets, the Frobenius norm remains relatively stable across layers but exhibits a slight peak in the middle and later layers. This stability suggests that DistilBERT maintains a balanced representation strength, avoiding overly high activations that could lead to unstable learning. The slight peak may indicate layers that contribute more significantly to information retention and transformation, especially on The Pile, where a higher Frobenius norm indicates potentially richer feature encoding.

The activation sparsity charts (rightmost in each row) show the proportion of zero activations per layer, offering insights into how sparse or dense the activations are. For both datasets, sparsity decreases towards the middle layers, followed by a slight increase in the final layers. This pattern suggests that early layers have sparse activations, possibly focusing on simpler, low-level features. In contrast, middle layers capture more complex representations, requiring more active neurons. The final layers exhibit slightly higher sparsity, which may reflect the model refining and focusing on specific features in its output.

In summary, this layer-wise analysis shows that DistilBERT processes data differently across The Pile and HackerNews datasets. The Pile dataset yields higher activation variance, L1 and L2 norms, and Frobenius norms, indicating more intense feature processing and possibly richer representations. In comparison, HackerNews maintains more balanced and consistent norms, suggesting that DistilBERT processes this data with less fluctuation across layers. These observations underscore the importance of layer-wise examination when evaluating model behavior across diverse datasets.

\section{Layer-wise Relevance Propagation (LRP) Evaluation Method}

Layer-wise Relevance Propagation (LRP) is an evaluation method that provides insight into the importance of each layer in a model by propagating relevance scores back through the layers. LRP is commonly used to understand which parts of the model contribute most significantly to its predictions. Here, we outline the mathematical foundation of the LRP process.

Given a neural network with layers indexed by \( l \) and a prediction function \( f(x) \), the goal of LRP is to assign a relevance score \( R_i^{(l)} \) to each neuron \( i \) in each layer \( l \). The relevance scores are initialized at the output layer with:

\begin{equation}
    R^{(L)} = f(x),
\end{equation}
where \( L \) is the final layer of the network and \( R^{(L)} \) represents the total relevance of the model's prediction.

LRP propagates relevance scores backward using a rule-based approach. One common rule is the \( \epsilon \)-rule, which distributes relevance scores based on neuron activations and weights, defined as:

\begin{equation}
    R_i^{(l)} = \sum_j \frac{a_i^{(l)} w_{ij}^{(l, l+1)}}{\sum_{i'} a_{i'}^{(l)} w_{i'j}^{(l, l+1)} + \epsilon} R_j^{(l+1)},
\end{equation}
where \( a_i^{(l)} \) is the activation of neuron \( i \) in layer \( l \), \( w_{ij}^{(l, l+1)} \) is the weight from neuron \( i \) in layer \( l \) to neuron \( j \) in layer \( l+1 \), and \( \epsilon \) is a small positive constant added for numerical stability.

Another common rule is the \( \alpha\)-\( \beta \)-rule, which divides relevance into positive and negative contributions. This rule is expressed as:

\begin{flalign}
    R_i^{(l)} = \sum_j  \alpha \frac{a_i^{(l)+} w_{ij}^{(l, l+1)+}}{\sum_{i'} a_{i'}^{(l)+} w_{i'j}^{(l, l+1)+}} \\
    - \beta \frac{a_i^{(l)-} w_{ij}^{(l, l+1)-}}{\sum_{i'} a_{i'}^{(l)-} w_{i'j}^{(l, l+1)-}}  R_j^{(l+1)} & \notag
\end{flalign}

where \( a_i^{(l)+} \) and \( a_i^{(l)-} \) represent positive and negative activations, \( w_{ij}^{(l, l+1)+} \) and \( w_{ij}^{(l, l+1)-} \) represent positive and negative weights, and \( \alpha \) and \( \beta \) are parameters that satisfy \( \alpha - \beta = 1 \).

The relevance scores are propagated through all layers until the input layer is reached, at which point each input feature \( x_k \) receives a relevance score \( R_k^{(1)} \):

\begin{equation}
    R_k^{(1)} = \sum_{j} \frac{x_k w_{kj}^{(0,1)}}{\sum_{k'} x_{k'} w_{k'j}^{(0,1)}} R_j^{(2)}.
\end{equation}

Finally, the sum of relevance scores across the input layer should ideally equal the model output:

\begin{equation}
    \sum_k R_k^{(1)} = R^{(L)} = f(x).
\end{equation}

This equality ensures that the relevance distribution is conserved, meaning the contribution from each input feature sums to the model's prediction score. LRP allows us to interpret which neurons and layers contribute most to the final output.

\section{Iterative Layer Pruning Process}

The iterative layer pruning process aims to reduce model complexity by removing layers with the least impact on model performance. The goal is to simplify the model while retaining its accuracy as much as possible. This section describes the pruning methodology mathematically.

Let \( \text{Performance}(f) \) represent the performance metric (e.g., accuracy) of a model \( f \). For each layer \( l \), we calculate a layer importance score \( I_l \), which quantifies the contribution of layer \( l \) to the model's performance. The importance score can be calculated using metrics like AVSS (Activation Variance-Sparsity Score) or EAVSS (Enhanced Activation Variance-Sparsity Score):

\begin{equation}
    I_l = \text{AVSS}_l.
\end{equation}

In each pruning iteration, we identify the layer \( l^* \) with the lowest importance score:

\begin{equation}
    l^* = \arg \min_l I_l.
\end{equation}

The layer \( l^* \) is removed from the model, creating a pruned model \( f' \). We then re-evaluate the model's performance with the remaining layers:

\begin{equation}
    \text{Performance}(f') = \text{evaluate}(f' | \text{data}).
\end{equation}

If the performance drop after pruning \( l^* \) exceeds a predefined threshold \( \delta \), the layer is retained; otherwise, it is permanently removed:

\begin{flalign}
    & \text{Remove Layer} \; l^* & \\
    & \text{if} \quad \text{Performance}(f') \geq \text{Performance}(f) - \delta. & \notag
\end{flalign}

To track the cumulative impact of pruning on model performance, we calculate the total performance loss after pruning \( n \) layers as:

\begin{flalign}
    \Delta \text{Performance}_{\text{total}} = \text{Performance}(f) \\
    - \text{Performance}(f^{(n)}) & \notag
\end{flalign}

where \( f^{(n)} \) is the model after pruning \( n \) layers. This metric helps to ensure that the cumulative performance loss remains within acceptable bounds.

An alternative approach to selecting \( \delta \) dynamically based on the overall model performance is to set \( \delta \) as a fraction of the initial model’s performance, such as:

\begin{equation}
    \delta = \alpha \times \text{Performance}(f),
\end{equation}
where \( \alpha \) is a scaling factor that determines the allowable percentage of performance loss per iteration.

The pruning process terminates when the relative performance difference between successive iterations falls below a small convergence criterion \( \epsilon \):

\begin{equation}
    |\text{Perform}(f^{(n)}) - \text{Perform}(f^{(n-1)})| < \epsilon.
\end{equation}

The final pruned model \( f^{\text{pruned}} \) has the following performance:

\begin{equation}
    \text{Performance}(f^{\text{pruned}}) \approx \text{Performance}(f),
\end{equation}
where \( f^{\text{pruned}} \) retains most of the original model’s accuracy but with fewer layers and reduced computational complexity.

This iterative pruning approach enables the creation of an efficient model by removing redundant layers while preserving its predictive power.

\end{document}